\documentclass{article}

\usepackage{arxiv}

\usepackage[utf8]{inputenc} 
\usepackage[T1]{fontenc}    
\usepackage{hyperref}       
\usepackage{url}            
\usepackage{booktabs}       
\usepackage{amsfonts}       
\usepackage{nicefrac}       
\usepackage{microtype}      
\usepackage{lipsum}		
\usepackage{graphicx}
\usepackage{natbib}
\usepackage{doi}

\usepackage{mathtools}
\usepackage{enumitem}
\usepackage{graphicx}
\usepackage[ruled,vlined,linesnumbered,noend]{algorithm2e}
\usepackage{subcaption}
\usepackage{xcolor}

\title{GLIDE-RL: Grounded Language Instruction through
DEmonstration in RL}

\author{ 
Chaitanya Kharyal \\ 
	Microsoft\\
	Hyderabad, India \\
	\texttt{chaitanyajee@gmail.com} \\
	\And
 Sai Krishna Gottipati \\
	AI Redefined Inc \\
	Montreal, Canada \\
	\texttt{sai@ai-r.com} \\
	\And
  Tanmay Kumar Sinha \\
	Microsoft  Research\\
	Bangalore, India \\
	\texttt{tanmaysinha18@gmail.com} \\
	\And
  Srijita Das \\
	University of Alberta \\
	Edmonton, Canada \\
	\texttt{srijita1@ualberta.ca} \\
	\And
   Matthew E. Taylor \\
	AI Redefined Inc \\
    University of Alberta \\
	Edmonton, Canada \\
	\texttt{matt@ai-r.com} \\
}

\hypersetup{
pdftitle={A template for the arxiv style},
pdfsubject={q-bio.NC, q-bio.QM},
pdfauthor={David S.~Hippocampus, Elias D.~Striatum},
pdfkeywords={First keyword, Second keyword, More},
}

\begin{document}
\maketitle

\begin{abstract}
One of the final frontiers in the development of complex human - AI collaborative systems is the ability of AI agents to comprehend the natural language and perform tasks accordingly. However, training efficient Reinforcement Learning (RL) agents grounded in natural language has been a long-standing challenge due to the complexity and ambiguity of the language and sparsity of the rewards, among other factors.  
Several advances in reinforcement learning, curriculum learning, continual learning, language models have independently contributed to effective training of grounded agents in various environments. Leveraging these developments,
we present a novel algorithm, Grounded Language Instruction through DEmonstration in RL (GLIDE-RL) 
that introduces a teacher-instructor-student curriculum learning framework for training an RL agent capable of following natural language instructions that can generalize to previously unseen language instructions. 
In this multi-agent framework,  the teacher and the student agents learn simultaneously based on the student's current skill level. We further demonstrate the necessity for training the student agent with not just one, but multiple teacher agents. Experiments on a complex sparse reward environment validates the effectiveness of our proposed approach.
\end{abstract}


\section{Introduction}

Grounded language learning can be defined as the task of learning the meaning of natural language units (e.g., utterances, phrases, or words) by leveraging the sensory data (e.g., an image) \cite{lang-groundinig-cirik-etal-2020-refer360}. It is a challenging task due to the inherent complexity, context sensitivity and ambiguity of natural language. This challenge is compounded when a decision-making agent has to perform a series of actions to complete different tasks expressed in natural language. Several works in the field of goal-conditioned RL~\cite{schaul2015universal,DBLP:conf/ijcai/LiuZ022} demonstrated the challenges associated with training goal-based RL agents in very sparse reward settings. These goals may be represented in many ways: for example, as one-hot vectors, as goal coordinates in a space (e.g., Euclidean), or as a `goal image' the agent needs to observe. They also share challenges with sparse reward RL tasks: credit assignment and sample efficiency. Representing goals in natural language can be useful because they are expressive and informative. Moreover, these goals can explicitly represent context sensitivity which is useful for learning policy in complex domains. 
However, representing goals in natural language also adds more complexity and ambiguity --- the agent needs to understand that ``grab the red ball'' is same as ``fetch that maroon sphere'' and it needs to perform a series of actions
before achieving an informative reward.
 
The above-mentioned challenges necessitate a framework involving curriculum learning \cite{2020JMLR} for the agent to learn anything useful and tame the beast i.e., the challenge of sparse rewards. We thus introduce a \textbf{teacher agent} that proposes goals to the \textbf{student agent}. Contrary to most other prior approaches, the teacher agent here also acts in the environment thus ensuring that all the goals it proposes are in fact reachable by another agent within a given time frame (e.g., episode length). Moreover, if the student fails to reach these goals, it can attempt to learn to clone the teacher's trajectory.
However, the bigger question remains: 
How is this setup useful for learning to follow natural language instructions? To address that, we introduce an \textbf{instructor agent} that attempts to describe the teacher's trajectory or key events in natural language and then converts them into the form of an instruction for the goal-conditioned student agent. This instructor agent can be of any form such as a pre-trained or a trainable video captioning module \cite{kuo2023mammut,xu2023mplug2} or a language model \cite{wang2023voyager,du2023guiding}.

In our work, we augment the environment with natural language descriptions of the events that the teacher had triggered and then convert them into an instruction format. 
We used a pre-trained language model (ChatGPT-3.5) to convert this single instruction to multiple synonymous instructions for the student agent to train on. This helped in evaluating and improving the generalization capabilities of the student agent. These interactions between the agents is summarized in Figure-\ref{fig:procedure} below.

\begin{figure*}[ht]
		\includegraphics[width=\textwidth]{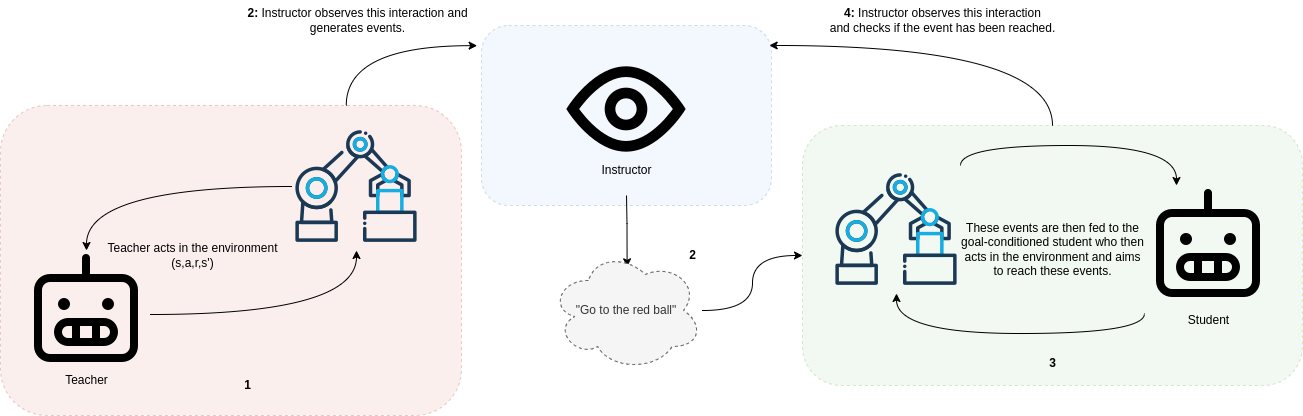}
		\caption{\small GLIDE-RL Algorithm has three independently functioning parts: \textit{The teacher, The instructor and The student}. (1) The teacher is an agent that acts in the environment to do complex things and gets rewarded based on the performance of the student. (2) The instructor observes these actions, describes them in the form of events and converts them to the instruction format. (3) The student is the goal-conditioned agent that strives to reach the goals set by the teacher, as described and instructed by the instructor. (4) The instructor also checks if the student has triggered the same/similar events in the environment to mark a certain state as success/failure for the student.}
    \label{fig:procedure}
\end{figure*}


\textbf{Contributions:} The key contributions of this work includes: 
\begin{enumerate}
    \item Introducing a novel algorithm (GLIDE-RL) and the Teacher-Instructor-Student framework for training RL agents grounded in natural language on sparse reward complex tasks
    \item Thorough empirical studies demonstrating the influence of factors like curriculum, behavior cloning, multiple teachers, type of language model on the performance of the Student agent
    \item Demonstrating generalization capabilities of the trained goal-conditioned RL agent across unseen goals and ambiguous instructions 
\end{enumerate}

\section{Related Work}
\label{sec:citations}

\noindent{\textbf{Goal conditioned RL:}} In goal-conditioned RL~\cite{schaul2015universal}, a representation of the goal that the agent needs to achieve is appended to the state representation such that the value function and policy is computed using this shared representation. Historical and key developments in the field of goal-conditioned RL are well summarized in~\cite{DBLP:conf/ijcai/LiuZ022,colas2022autotelic} 
In attempts to  make the training more efficient, several algorithms were proposed that suggest improvements over the way the goals were generated and in the overall training process. In Hindsight experience replay~\cite{andrychowicz2017hindsight}, the target goal set was expanded with the intermediate goals that the agent actually achieved while trying to reach the original target goal. Florensa et al.~\cite{florensa2018automatic} used a GAN setup where the generator generates goals of appropriate difficulty for the RL agent and the discriminator identifies whether the generated goals belong to the goal space of intermediate difficulty. Campero et al.~\cite{campero2020learning} proposed a teacher-student framework where the teacher generates curriculum of increasingly difficult goals based on the student's skill level and the student gets rewarded implicitly on reaching the proposed goals. 

Asymmetric Self Play (ASP)~\cite{sukhbaatar2018intrinsic} is an unsupervised training approach where the teacher agent proposes increasingly challenging goals for the student to achieve by exploring the environment. The teacher and the student  are both RL agents; where the former gets rewarded for suggesting goals outside the student's comfort and the latter gets rewarded for achieving the proposed goal. This was later extended by Plappert et al.~\cite{openai2021asymmetric} for complex robotics task and Du et al.~\cite{du2022takes} for more feasible and challenging goals using gameplay between two teachers and two students. Kharyal et al.~\cite{kharyal2023you} extended ASP with multiple teachers that suggested diverse goals to help the student agent learn faster. All the prior work condition the goals in the same space as that of the agent. Our work incorporates natural language goals in the goal-conditioned RL agent's representation in addition to introducing the Instructor agent; thus taking a step forward towards the ability of humans to convey denser knowledge to these agents.

\noindent{\textbf{Grounded language learning:}} Recently, there have been a lot of efforts to integrate natural language instructions for Reinforcement Learning. Luketina et al.~\cite{luketina2019survey} categorize these directions as language conditional RL where agent-language interaction is part of the problem formulation \
and language assisted RL where natural language is used to guide the training of RL agents. There are a few works where the agent follows instructions provided to it in the form of a high-level goal or a policy description~\cite{bahdanau2018learning,hermann2017grounded,chaplot2018gated} in natural language. Oh et al.~\cite{oh2017zero} proposed an approach to train on a shorter sequence of instructions and generalize to a longer sequence of instructions (seen and unseen) by learning parameterized skills and using them to execute the relevant instructions. However, this includes policy pre-training to understand the structure of these instructions. Goyal et al.~\cite{goyal2019using} and Wang et al.~\cite{wang2019reinforced} used the agreement between trajectory and expert provided language instructions for reward shaping in sparse reward tasks. Chan et al.~\cite{chan2019actrce} combines Hindsight experience replay~\cite{andrychowicz2017hindsight} with natural language goal description which is provided by a teacher based on what the RL agent does in the specific episode.  Du et al.~\cite{du2022takes} takes a step further and queries pretrained large language models to get context-sensitive and diverse goal descriptions. Our work is aligned with these advancements in grounded language in RL but differs in the way the goals are generated and communicated --- The Teacher agent acts in the environment to ensure that the goals generated are infact reachable and within the zone of proximal development~\cite{seita2019zpd} of the language-conditioned agent. Moreover, the Instructor agent generates a database of synonymous instructions to ensure that the language-conditioned agent generalizes to unseen instructions.

\section{Background}{\label{sec:background}}
\noindent \textbf{Reinforcement Learning:} The teacher and the student agents operate on the framework of a markov decision process (MDP). An MDP consists of a five-element tuple $(S, A, P, R, \gamma)$ where $S$ is the set of all possible states, $A$ is the set of all possible actions, $P$ is the transition probability function, $R$ is the reward function, and $\gamma$ is the discount factor. At each time step t, the agent is at a state $s_t \in S$ and takes an action $a_t \in A$. The environment dynamics map the state-action pair into a successor state $s_{t+1} \sim P(\cdot|s_t, a_t)$, and the agent receives a scalar reward $r_t \sim R(s_t, a_t)$. In some environments, the agents might not have access to the full state of the environment. In such cases, a partial observation $o_t$ is used as input to the RL agent. The objective is to find a policy $\pi : S \rightarrow A$ that maximizes the expected discounted return $\mathbb{E}[\Sigma_{t=0}^{\infty} \gamma^{t} r_t]$ with $\gamma \in (0, 1]$. The teacher agents are trained using this simple MDP formulation whereas the student agent is trained in goal-conditioned setting. 

Goal conditioned RL deals with training agents to reach the goals determined at the beginning of an episode by the environment or an external agent (like the teacher agent in our context). In these settings, the agent also takes in the goal $g$ as input at every time step in addition to the observation $o_t$. All the RL algorithms used in a standard MDP above can also be used for this goal-conditioned MDP. The goal $g$ in our experiments is a natural language instruction.

\noindent \textbf{Deep Q-Networks:} The action space is discrete in our experiments. We thus chose D3QN algorithm \cite{d3qn} for all the experiments. It consists of 'double' \cite{double-dqn} and 'dueling' \cite{dueling-dqn} techniques added to the DQN algorithm \cite{dqn}. Note that both the teacher and student agents have the same action space in our setting. They primarily differ in their reward structure (which we design to be adversarial in nature). The input to the student agent also includes the natural language instruction in addition to the usual input observation. 

A typical DQN loss \cite{dqn} is given by: 
\begin{equation}
L_i\left(\theta_i\right)=\mathbb{E}_{s, a, r, s^{\prime}}\left[\left(y_i^{D Q N}-Q\left(s, a ; \theta_i\right)\right)^2\right]
\label{eq:dqnloss}
\end{equation}

where, 
$y_i^{D Q N}=r+\gamma \max _{a^{\prime}} Q\left(s^{\prime}, a^{\prime} ; \theta^{-}\right)$ is the temporal difference (TD) target and $\theta^{-}$ represents the parameters of a fixed target network. 
To avoid the problem of overestimation of Q-values, double DQN \cite{double-dqn} uses the following TD target:
\begin{equation}
y_i^{D D Q N}=r+\gamma Q\left(s^{\prime}, \underset{a^{\prime}}{\arg \max } Q\left(s^{\prime}, a^{\prime} ; \theta_i\right) ; \theta^{-}\right)
\label{eq:ddqntarget}
\end{equation}

In dueling DQN \cite{dueling-dqn}, two network heads are used, one for value function $V(s; \theta, \beta)$ and one for advantage function $A(s, a^\prime, \theta, \alpha)$. Note that $\theta$ are the shared parameters. The Q-function is then estimated as:
\begin{equation}
\begin{aligned}
& Q(s, a ; \theta, \alpha, \beta)=V(s ; \theta, \beta)+ \\
& \qquad\left(A(s, a ; \theta, \alpha)-\frac{1}{|\mathcal{A}|} \sum_{a^{\prime}} A\left(s, a^{\prime} ; \theta, \alpha\right)\right)
\label{eq:d3qnadv}
\end{aligned}
\end{equation}

In D3QN \cite{d3qn}, both the double and dueling techniques are combined and has been proven to work well on a wide range of environments with discrete action spaces. We thus use it as a base agent in all our experiments. Both the teacher and student agents are instances of the D3QN actor class. We also use the frame stacking technique \cite{dqn} to assist in learning.

\section{Proposed Algorithm}
\label{sec:methods}

The final objective is to have an RL agent capable of following the natural language instructions in a simulated environment with sparse reward. In our proposed framework \textbf{G}rounded \textbf{L}anguage \textbf{I}nstruction through
\textbf{DE}monstration in \textbf{RL} (GLIDE-RL), this agent is designated as the Student agent.

We have three types of agents: Teacher, Instructor and Student. The teacher and student agents are trained in an adversarial setup. While the student is a goal-conditioned agent that aims to complete the tasks provided to it as natural language instructions, teachers are trained to propose tasks/goals by acting in the environment that student agent can't achieve --- this results in teachers providing a curriculum  of incrementally harder goals for the student agent to train on. We train multiple teacher agents to assist in better generalisation of the student agent by proposing diverse goals. Note that the Teacher agents by themselves are not capable of describing what they have done or to instruct the agents. It is the role of the instructor agent to describe what the teacher has done in natural language and then convert it to a form of instruction for the student agent to act and train upon.
\noindent Formalizing our problem set-up as below:\\
\noindent {\bf Given:} A student agent S, a set of teacher agents $\{T_1,T_2,\cdots,T_N\}$ and an instructor agent $I$\\
\noindent {\bf To-do:} Learn an optimal goal-conditioned policy $\pi_S$ for the student agent that can follow natural language instructions generated by $I$ by observing the evolving teacher policies $\{\pi_{T_1},\pi_{T_2},\cdots,\pi_{T_N}\}$ \\
\noindent {\bf Assumptions:} We make the following assumptions (1) All the teacher agents start from scratch with a random policy and only learn from feedback (reward) related to the student's performance (2) Instructor agent is capable of describing the actions of the teacher in natural language and is equipped with a pre-trained LLM to convert these descriptions to several synonymous instructions.

GLIDE-RL is detailed in Algorithm~\ref{algo:proposal} and illustrated in Figure-\ref{fig:procedure}. We denote the $n$ teacher agents as $T_1$, $T_2$,... $T_N$ and the student agent as $S$. We represent the parameters of the teacher's networks with $\theta_{T_1}, \cdots, \theta_{T_N}$ and that of the student agent with $\theta_S$. We denote the parameters of the language model with $\phi_L$ and that of the instructor with $\phi_I$.

In every student-teacher rollout, one of the teachers $T_i$ acts in the environment by choosing an action $a_t$ according to its policy $\pi_{T_i}$ based on the current observation $o_t$ until the end of episode (of predefined length). The state-action ($s_t$, $a_t$) pairs that the teacher encountered in its trajectory are then used by the Instructor to describe in natural language the course of events that the teacher has triggered. In one rollout, the teacher could trigger multiple events $E_i$. The instructor first describes these events $E_i$ in natural language (e.g., "you are standing in-front of red ball") and then converts this description to the form of an instruction (e.g., "go to the red ball").

The instructor then uses a pre-trained language model $\phi_I$ to generate $m$ synonymous instructions $I_{i1}, I_{i2}, ... I_{im}$. The first subscript $i$ indicates that it's an instruction corresponding to the event $i$ whereas the second subscript $0, 1, 2, ... m$ just denotes the $j^{th}$ synonymous instruction for the same event $i$. These events then become the goal for the student agent. Events are fed to the student agent one at a time, in the exact same order $E_1$, $E_2$, ....  $E_n$ as the teacher has triggered them, in the form of natural language instructions. Thus, in addition to its current observation $o_t$, the goal-conditioned student agent also takes in randomly sampled task/goal $I_{ij}$ starting with $i = 0$ and $j \in [0, m]$. the language model $\phi_L$ transforms the natural language instruction to an embedding (a tensor) which is then concatenated with the input observation. This combined input representation is then passed through Deep Neural Network to obtain the Q-values for every action. The action corresponding to the maximum Q-value is chosen.the language model transforms the natural language instruction to an embedding (a tensor) which is then concatenated with the input observation. 
At each time step, the student gets 0 reward if it doesn't finish the task/goal $E_i$. If it finishes, it gets a positive reward (this value is hyperparameter tuned) and, from the next time step it takes in task/goal $E_{i+1}$ as input in addition to the observation input $o_t$ and strives to achieve the goal $E_{i+1}$. 

The student continues to act in the environment until it finishes all the goals $E_1$, $E_2$, ... $E_n$ or until the maximum episode length. Note that the number of events $n$ a teacher will trigger can vary across rollouts. An important implementation detail to note is that the done flag is marked as true for the Student after it finishes each individual event $E_i$. During the training process, this ensures that the Student is rewarded to focus only on the immediate goal given its current observation and the goal it is conditioned on.

We then update both the student and teacher networks using the standard D3QN loss described earlier in Equations-\ref{eq:ddqntarget} and \ref{eq:d3qnadv} of Section-\ref{sec:background}. We denote this loss with $\mathcal{L}_{D3QN}$. We also employ Behaviour Cloning (BC) loss to train the student to tackle the sparse reward nature of the problem and assist the student's learning by using the teachers' behaviours similar to ~\cite{openai2021asymmetric}. The BC loss we use is defined as follows:
\begin{equation}
\mathcal{L}_{BC} = - \mathbb{E}_{(s_t, g_t, a_t) \sim D_{BC}} \left[ \log \frac{e^{\pi_S(a_t | s_t, g_t)}}{\sum_i e^{\pi_S(i | s_t, g_t)}} \right]
\end{equation}
Where $s_t$, $a_t$ and $g_t$ are the teacher's state, the action it took in that state, and the first event it triggered after performing the action respectively;$\pi_S$ is the student's policy. 

$D_{BC}$ is the behavioural cloning replay buffer, constructed to help the Student learn to trigger the events that it failed to trigger during the roll-out
Therefore, we only insert those (event, state, action) tuples that the student is not able to trigger, thereby not confusing the student about previously learnt goals.

Therefore, the overall loss function for the student becomes:

\begin{equation}
\mathcal{L} = \mathcal{L}_{D3QN} + \Gamma\mathcal{L}_{BC}
\end{equation}

Where $\Gamma$ is the adaptive behavioural loss coefficient which is calculated as follows on every update:

\begin{equation}
    \Gamma_{t+1} = \Gamma_{t} + (\alpha \mathcal{L}_{RL} - \mathcal{L}_{BC})\epsilon
\end{equation}

Where $\alpha$ is a predefined constant called \textit{BCL ratio} and $\epsilon$ is the predefined decay rate for $\Gamma$. Every teacher and student agent uses its own rollout data (experience) to update their respective parameters. 

\textbf{The key advantages of this training setup} are two-fold --- 
(1) we know by construction, that the events triggered by a teacher are reachable by the student from the starting state within episode length (2) the teacher provides a valid demonstration on how the events can be reached, which can be leveraged using the Behavioural Cloning Loss for goals that the student fails to reach.

\textbf{Advantage of using multiple teachers} is that these teachers learn diverse policies due to the way we structured the reward function --- A teacher gets a negative (positive) reward if the student is able (unable) to reach the goal. This implies that the teachers are incentivized to set digoals different from each other because the student agent is being trained to reach the goals that are already set by other teachers and can reach the same goal again if the current teacher sets the same goal. Therefore, each teacher's demonstrations are intrinsically influenced by other  teachers' demonstrations.

\textbf{Environment configuration:} While training the agents, we reset the environment to a random configuration at the beginning of every roll-out. Within a roll-out, environment configuration remains the same. This ensures that the teacher and student have the same initial conditions which enables a teacher to set a meaningful goal and for the student to learn useful behaviors from cloning the teacher's trajectory.

\begin{algorithm}[!h]
\caption{GLIDE-RL Algorithm}
\SetAlgoLined
\SetKwComment{Comment}{//}{ }
\label{algo:proposal}
\KwData{$N$ \Comment*[r]{Number of teacher agents}} 
\KwData{$\theta_{T_1}, \cdots, \theta_{T_N} , \theta_S$ \Comment*[r]{Parameters for the agents}}
\KwData{$\phi_L, \phi_I$ \Comment*[r]{Parameters for Language model and Instructor}}
\For{\texttt{rollout} = $1,2,\cdots$}{
    $k = \texttt{rollout}//N $\;
    \texttt{teacher} = $k^{th}$ \texttt{teacher}\;
    \texttt{goal = $[]$}\;
    \Comment*[l]{run teacher}
    \For{\texttt{episode length}}{
        \texttt{state} = \texttt{teacher.act(state)}\;
        \texttt{event} = \texttt{Instructor.describe()}\;
        \If{\texttt{event} != ""}{
            \texttt{goal.append(event)}\;
        }
    }
    \Comment*[l]{run student}
    \For{\texttt{episode length}}{
        \texttt{g} = \texttt{language\_model(goal[0])}\;
        \texttt{state} = \texttt{student.act(state, g)}\;
        \If{\texttt{Instructor.reached(goal[0])}}{
            \texttt{goal.pop(0)}
        }
    }

    \texttt{Distribute rewards}\;
    \texttt{Update agents using RL and BC losses}\;  
}
\end{algorithm}
\textbf{Reward structure:}
The teacher gets a reward of $+y$ if the student fails to trigger the event $E_i$ and a reward of $-x$ if the student reaches the event. Additionally, the teacher also gets rewarded $-C$ if it fails to trigger any event throughout the episode. On the other hand, the student gets rewarded $+z$ if it is able to trigger the goal event and $0$ if it fails to do so. . We experimented with different reward values for each $n$-teacher setting, where $n \in \{1, 2, 4\}$ More details are shared in Section-\ref{sec:result}. Note that, while the reward structure is adversarial in nature, it is not a zero-sum setup.

\begin{equation*}
          R_T = \begin{cases*} 
           -x & if student reaches $E_i$\\
           +y & if student does not reach $E_i$\\
           -C & if the teacher doesn't trigger any event throughout \\  & the episode
       \end{cases*}
    \end{equation*}
    
\begin{equation*}
          R_S = \begin{cases*} 
           +z & if student reaches $E_i$\\
           0 & if student does not reach $E_i$
   \end{cases*}
\end{equation*}

where $\{x,y,z,C\} > 0$ and $C > x,y$.

Furthermore, to encourage teacher's exploration, we give it an additional reward for triggering new events. This reward decays with the number of times the event has been triggered as:

\begin{equation*}
    R_{T}^{ex}(E_i) = 3 * 0.97^{f_{E_i}}
\end{equation*}

where $f_{E_i}$ is the frequency of event $E_i$  triggered by any teacher.

\section{Experimental Results}
\label{sec:result}

\subsection{Experiment Setting - BabyAI}

\begin{figure}[htp]
    \centering	\includegraphics[width=0.5\linewidth]{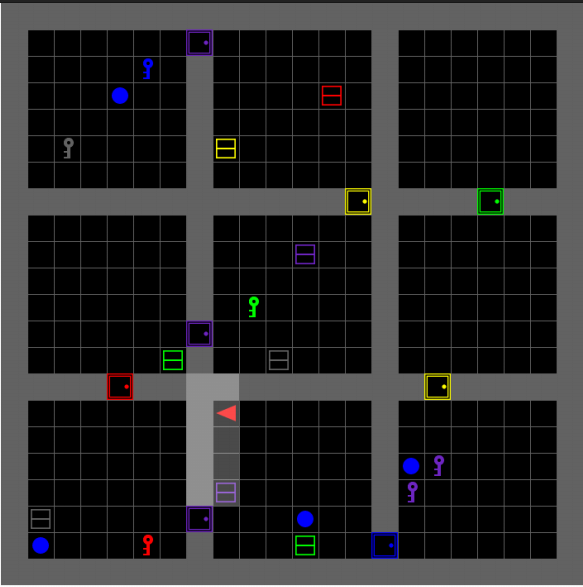}
		\caption{\small BabyAI BossLevel environment}
    \label{fig:bosslevel}
\end{figure}

\textbf{Environment:}
We use 'BossLevel' environment, the most complex environment within the babyAI suite \cite{babyai} for all our experiments. As shown in the Figure-\ref{fig:bosslevel}, the environment consists of 9 rooms with each room having the size of 6x6. There are different types of objects: \{ball, box, key, door\}. Each object can be of any color: \{red, green, blue, grey, purple, yellow\}.     
At the beginning of every episode, the agent along with all objects are spawned in random positions. 

The observation to the agent is a $(7\times 7 \times 3)$ image. This image has encoding for the different objects, their colours and states in different layers. The action is discrete in nature and selects one of the following actions (\emph{turn left, turn right, move forward, pickup, drop, toggle, done)}.

\noindent \textbf{Test set:} 
For generating the test set, we let a random agent act in the environment for 1000 timesteps (as compared to the student's episode length of only 115 timesteps). We let it run for 100 episodes and store all the events triggered by this random agent. We also store the environment's initial states (including agents  initial positions). While testing, the environment is initialized with these initial states and the goal-conditioned student agent is instructed to trigger these events in the exact same order as the random agent in all the 100 episodes. One test '
goal' includes multiple events (tasks) and the student's episode is considered as successful only if it triggers all these events in the exact same sequence that the random agent had triggered them. 

\noindent\textbf{
Instructor:}
To enable better generalization of the student agent, we convert the event description to an instruction and generate a set of 50 synonymous instructions for each event. An example of a synonymous instruction for the event "standing infront of the red ball" (described by the environment) can be "lift up the crimson ball". During training, we randomly sample an instruction from the corresponding synonymous instruction set for each event triggered by a teacher agent. For the experiments that involve testing the generalization capabilities of the agent, we keep 5 instructions from each synonymous instruction set, in a holdout test set. These instructions are not used during the training and are only used as part of the test set.

For converting these natural language instructions into language embeddings, we use off-the-shelf language models. We experimented with \textit{all-distilroberta-v1}, \textit{all-mpnet-base-v2} and \textit{multi-qa-mpnet-base-dot-v1} models provided by the \textit{sentence\_transformers} package in \textit{python}.
We noticed that \textit{all-distilroberta-v1} performed consistently better than other language models (as observed in hyperparameter sweeps) and used it for all the experimental results reported in this paper.

\begin{table}[htp]
  \caption{Hyperparameters used for D3QN for 4-teacher run}
  \label{tab:hyp_4-teach}
  \begin{tabular}{lll}\toprule
    \textit{Hyperparameter} & \textit{Value} \\ \midrule
    BCL ratio ($\alpha$) & 0.900 \\
    Frame stack & 8 \\
    gradient clip & 0.67 \\
    Optimizer & Adam\cite{kingma2017adam} \\
    Learning rate & $5.13\times10^{-5}$ \\
    $\tau$ & 0.098 \\
    Student reward ($z$) & 3 \\
    Teacher reward ($C$) & 8 \\
    Teacher reward ($y$) & 6 \\
    Teacher reward ($x$) & 2 \\
    \bottomrule
  \end{tabular}
\end{table}

\subsection{Results and Analysis}
In this section, we empirically demonstrate the effectiveness of our proposed algorithm.
The evaluation of our algorithm aims to answer the following research questions:
\begin{enumerate}[wide, labelindent=0pt, nosep]
    \item[\textbf{R1:}] How good is the performance of GLIDE-RL on events seen during training?
     \item[\textbf{R2:}] How effective is our algorithm in understanding synonymous goals/instructions seen during training? 
    \item[\textbf{R3:}] Does the student agent learn better with increasing number of teachers? 
    \item[\textbf{R4:}] Is the student able to generalize to previously unseen goals?
\end{enumerate}

\noindent \textbf{Experiment 1: Ability to succeed on events seen during the training}

In this experiment, we aim to test if the student is able to succeed on the synonymous instructions seen during training. We also attempt to understand the effect of various components like having a curriculum and behaviour cloning loss on the training.

It is important to note that while these instructions are seen during training, the actions the Student agent needs to take to finish that instruction are different --- the positions of different objects and the agent's starting position are generated randomly for the test set and these are not used during training. The sequence of events that the Student must accomplish are also not seen during the training.

\begin{figure}[htp]
    \centering	\includegraphics[width=\linewidth]{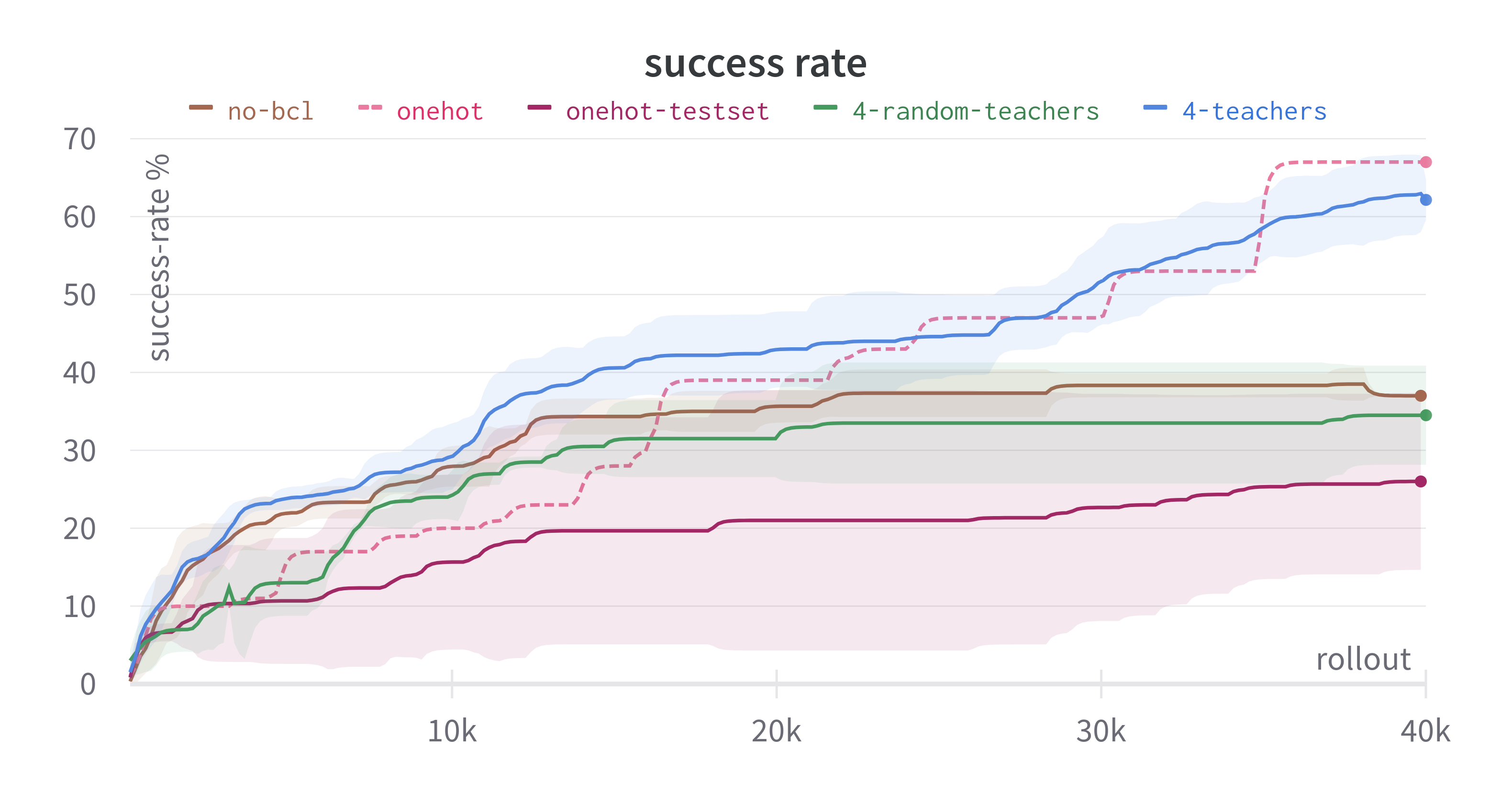}
		\caption{\small Success rate of different variations on the test set. The plot shows the mean and standard deviation over 5 seeds}
    \label{fig:ex1}
\end{figure} 

We train various agents (and establish baselines) to show the importance of teachers (and the curriculum generated by these teachers) for the training of the student:

Firstly, we train a student conditioned on one-hot goals (onehot in figure \ref{fig:ex1}). Teachers' functionality doesn't change here. But, for the student, instead of receiving language embeddings from a language model as inputs, it receives pre-designed one-hot encoding for each event. There is no notion of synonymous goals here, the events triggered by the Teacher are directly converted to a one-hot encoding and sent to the agent. This baseline gives us an estimate of the upper bound of success rate achievable. This also establishes that not all the event sequences of the test set can be completed in just 115 time steps that took the random agent 1000 time steps. Note that we only allow 115 timesteps for the student agent to ensure that the task remains challenging. Also note that, with one-hot encodings, the agent does not have any generalization capabilities as the size of the encodings cannot be increased to accommodate the unseen goals.

With the upper bound on the performance (on the test-set) being established, we now train the student conditioned on the embeddings from the language model (4-teachers in figure \ref{fig:ex1}) 
Here, we present the student with the synonymous events while training as described before. The aim of this is to gauge how well can the student perform on the test set as compared to when trained with one-hot encoding.

To show the importance of the curriculum generated by the teachers, we train a student with random teacher agents (4-random-teachers in figure \ref{fig:ex1}), and another one-hot student trained directly on the test set but without any teachers or curriculum (onehot-testset in figure \ref{fig:ex1}). While the random-teachers don't learn adversarially with the student, and hence provide no curriculum, the onehot-testset baseline doesn't have the notion of teachers. We introduced onehot-testset baseline to understand how challenging the task is without a curriculum set by the teachers.

Furthermore, to understand the necessity of Behavioral Cloning Loss (BCL), we train another baseline (no-bcl in figure \ref{fig:ex1}). We train this student in the exact same manner as GLIDE-RL, with the exception of not using the BCL while training.

Figure-\ref{fig:ex1}
shows the importance of teachers (and curriculum) in general for the student's performance as the students trained without the teachers' curriculum fail to perform well (measured in terms of success rate), even when trained directly on the test set.
Furthermore, we see that the student with goals conditioned as language embeddings is able to perform comparable to the one with one-hot goals. 
Moreover, with the no-bcl baseline, we establish the importance of the behavioural cloning loss during the training process --- without BCL, the student's performance is as good as the scenario with random teachers (no curriculum).

\noindent \textbf{Experiment 2: Ability to generalise on synonyms seen during training}

\begin{figure}[h]
  \centering
  \begin{subfigure}[b]{0.4\textwidth}
    \includegraphics[width=\linewidth]{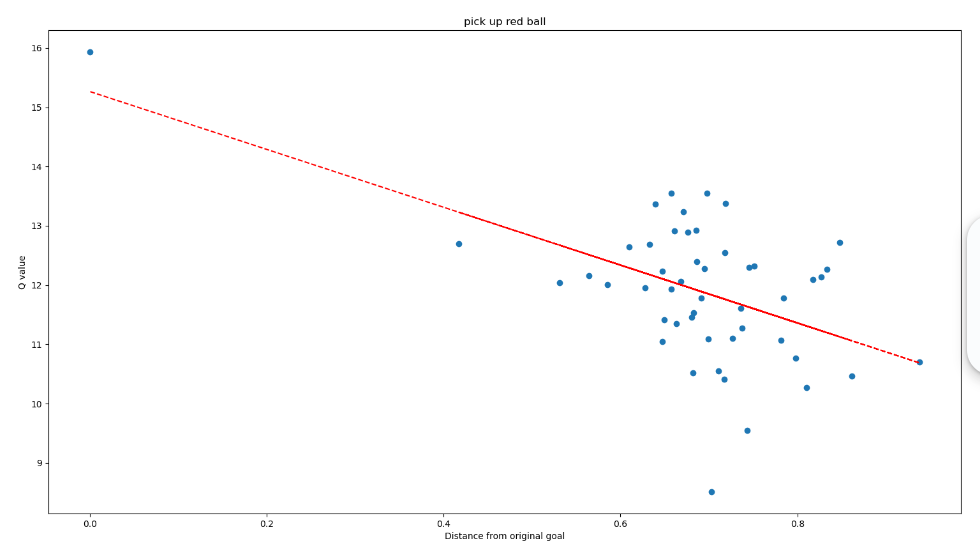}
  \end{subfigure}
  \begin{subfigure}[b]{0.4\textwidth}
    \includegraphics[width=\linewidth]{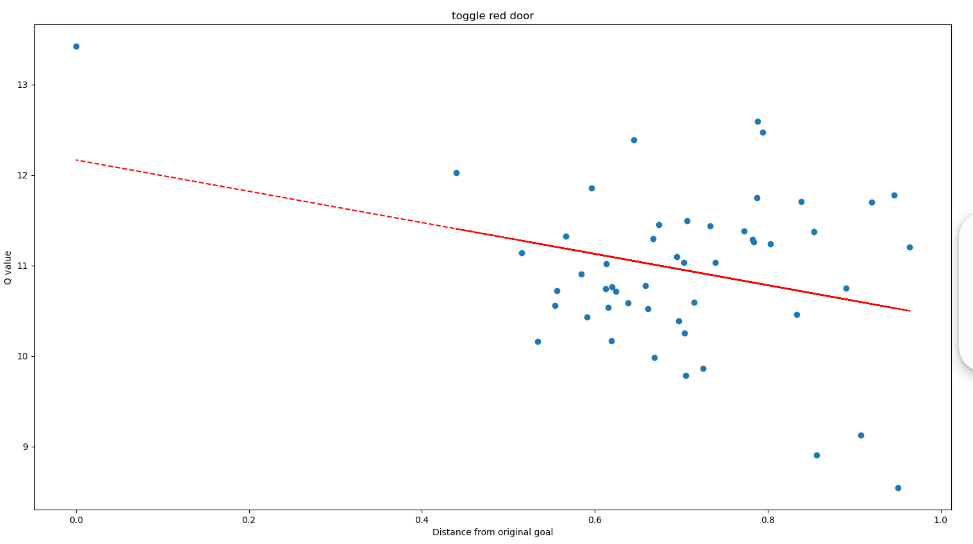}
  \end{subfigure}
  \caption{Q values of the synonymous events vs distance from the original event. Each blue dot (\textcolor{blue}{$\bullet$}) represents one synonymous event for the original event. The \textcolor{red}{red} line represents the trend among the points}
  \label{fig:qvsdist}
\end{figure}

Here, we try to reason why the student might be able to accomplish some of the instruction synonyms better than others. We hypothesise that some synonyms of different instructions (corresponding to different events) overlap in the embedding space, thus making it difficult for the student to generalise on those events.

To test this hypothesis, we run the trained Student agent on the test set and for each event in the test set, just before the student triggers the event, we check the maximum Q value in that state corresponding to each of the synonyms. Finally, we average the Q value for each synonym over the occurrence of the event. 

We notice an interesting, yet expected, pattern when we plot Q values of various synonyms for an event versus the distance from that event in the embedding space (fig: \ref{fig:qvsdist}). We see that as the distance of a synonymous instruction from the original instruction (which should be close to the centre of the synonyms cluster in the embedding space because all the synonyms are written using that as the root event) increases, the Q value decreases.

\noindent \textbf{Experiment 3: Effect of varying the number of teachers}

\begin{figure}[htp]
    \centering	\includegraphics[width=\linewidth]{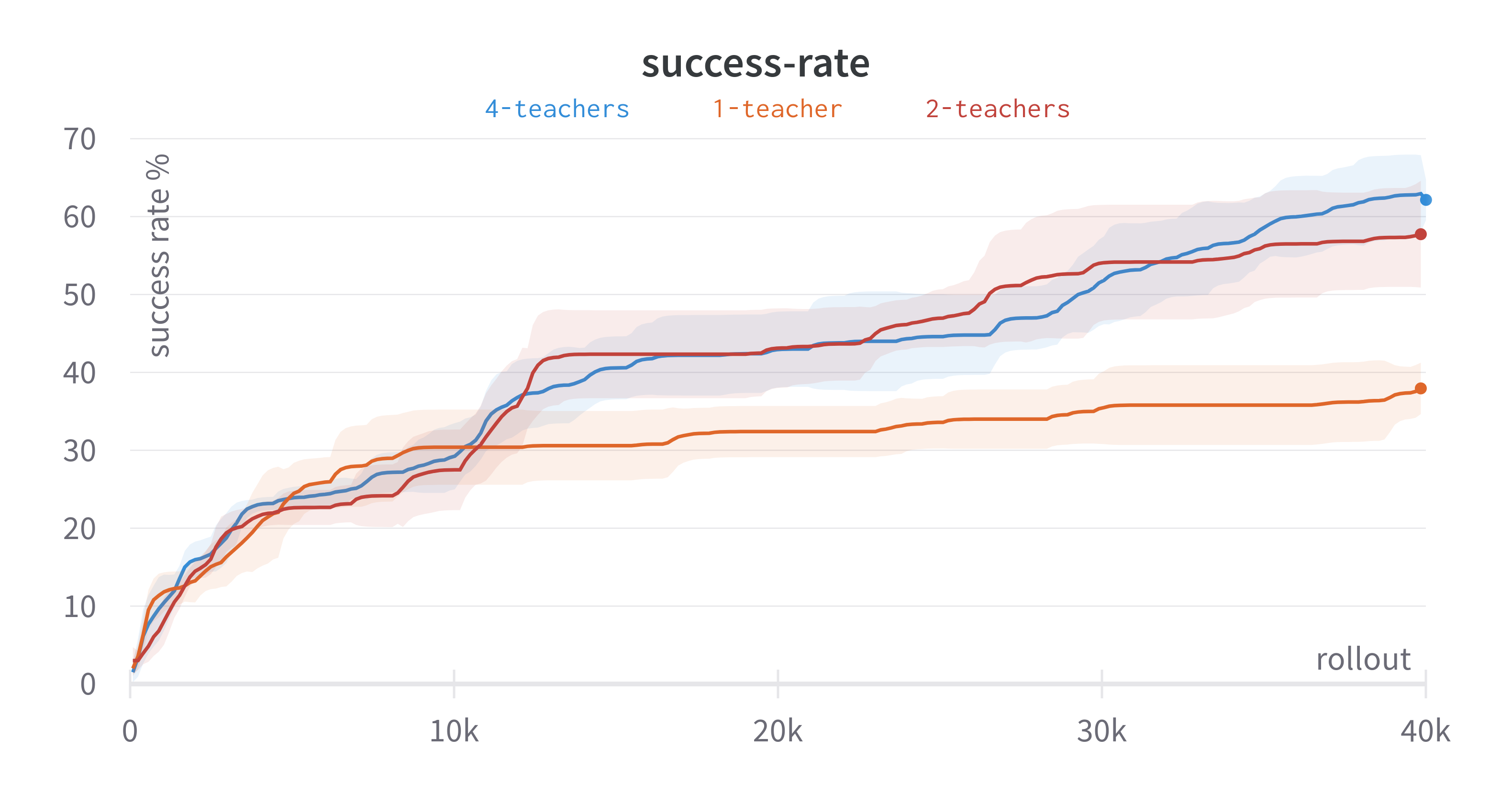}
		\caption{\small Success rate on the test set with varying number of teachers. The plot shows mean and standard deviation over 5 seeds}
    \label{fig:ex3}
\end{figure} 

In this experiment, we try to understand how changing the number of teachers affects the success rate of the student. Similar to previous experiments, each curve in Figure-\ref{fig:ex3} is averaged over 5 seeds. We notice that the success-rate on the test-set increases as we increase the number of teachers. We believe that this is because of the diversity of the goals being generated by different teachers as established by Kharyal et al \cite{kharyal2023you} in a similar experimental setup, but with simpler goal representation (Euclidean coordinates) 
and on much simpler environments. Further experimentation with more number of teachers is an avenue for exploration in future work. Given the complexity of the environment, and considering that the 4-teacher scenario is only slightly better than the 2-teacher scenario, we could hypothesize that the success-rate would only increase marginally as we increase the number of teachers further.

\noindent \textbf{Experiment 4: Ability to generalise on previously unseen goals}

\begin{figure}[htp]
    \centering	\includegraphics[width=\linewidth]{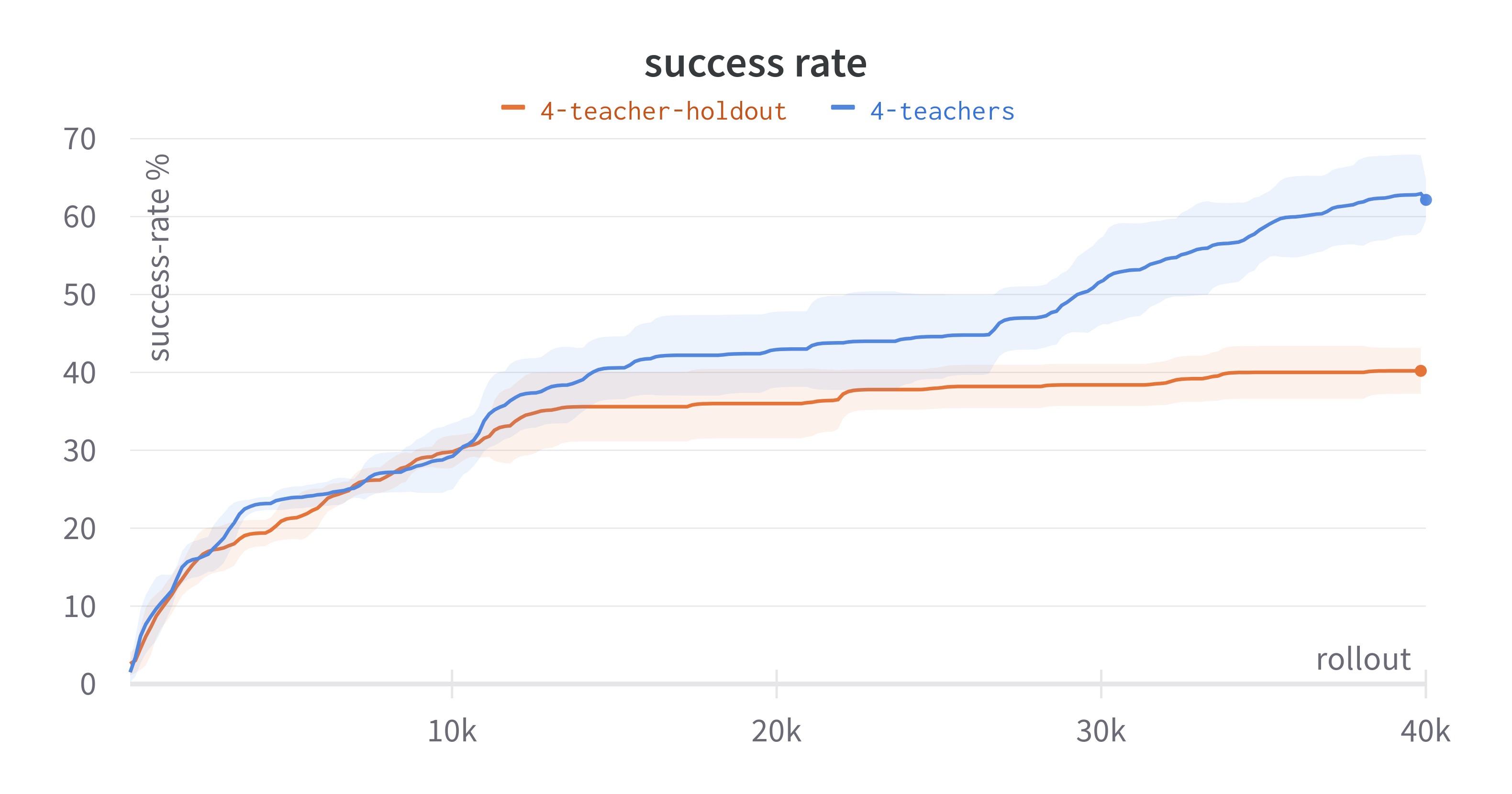}
		\caption{\small Success rate of the student on holdout-test-set}
    \label{fig:holdoutsuccess}
\end{figure}

To test the ability of the student agent to generalise on instructions unseen during training, we construct a holdout set of 5 synonymous instructions for each event and hold them out of the training process. Then, we test the student agent on goals constructed purely using the sequence of events corresponding to these holdout set of instructions denoted as 4-teacher-holdout in Figure-\ref{fig:holdoutsuccess}. We compare the performance of this agent with the agent that was trained on the complete instruction set (including the holdout test set denoted as 4-teacher). Both these student agents were trained with four teachers (as that was the best performing setup as noted in Figure-\ref{fig:ex3}). 

We can observe in Figure-\ref{fig:holdoutsuccess} that 4-teacher-holdout variant slightly lags behind the performance of the agent that was trained on the complete instruction set (4-teacher) throughout the training process. However, it is important to note that 40 percent success-rate is still a remarkable performance given that the agent has never seen any of the 1) sequence of events or 2) language instructions or 3) the positions of the agent or objects in the environment during its training process. 

\section{Conclusion and Future Work}

In this work, we proposed a novel algorithm and framework for training RL agents grounded in natural language and demonstrated its ability to generalize to unseen language instructions. We have also shown the impact of language model and the number of teachers on the performance of the student agent. We would like to extend this study to include the training of the instructor agent as well in more complex environments. Another possible direction is using actual humans as instructors for the proposed framework. We would also want to explore the different kinds of skills that the teacher agents learn in the process of challenging the student agent.

\label{sec:limitations}




\bibliographystyle{unsrtnat}

\end{document}